%% file: paper.tex
\documentclass[11pt,a4paper]{article}
\usepackage[hyperref]{acl2018}
\usepackage{times}
\usepackage{latexsym}
\usepackage{booktabs}
\usepackage{graphicx,url,color}
\usepackage{amsmath,amssymb,amsopn,amsthm,bm,bbm}
\usepackage{algorithm,algpseudocode,algorithmicx}
\usepackage[textsize=small]{todonotes}
\usepackage{verbatim}
\usepackage{multirow,rotating}
\usepackage[normalem]{ulem}
\usepackage{placeins}
\usepackage[procnames]{listings}

\usepackage{url}

\aclfinalcopy 
\def\aclpaperid{723} 

\newcommand{\D}{\mathcal{D}}

\title{
	Improving a Neural Semantic Parser by Counterfactual Learning from Human Bandit Feedback
}

\author{Carolin Lawrence\\
	Computational Linguistics\\
	Heidelberg University\\
	69120 Heidelberg, Germany\\
	{\tt \small Lawrence@cl.uni-heidelberg.de}
	\And
	Stefan Riezler\\
	Computational Linguistics \& IWR\\
	Heidelberg University\\
	69120 Heidelberg, Germany\\
	{\tt \small riezler@cl.uni-heidelberg.de}}
\date{}

\begin{document}
	\maketitle
	\begin{abstract}
		Counterfactual learning from human bandit feedback describes a scenario where user feedback on the quality of outputs of a historic system is logged and used to improve a target system. We show how to apply this learning framework to neural semantic parsing. From a machine learning perspective, the key challenge lies in a proper reweighting of the estimator so as to avoid known degeneracies in counterfactual learning, while still being applicable to stochastic gradient optimization. To conduct experiments with human users, we devise an easy-to-use interface to collect human feedback on semantic parses. Our work is the first to show that semantic parsers can be improved significantly by counterfactual learning from logged human feedback data.  
	\end{abstract}
	
	\input{sec-intro.tex}
	\input{sec-related.tex}
	\input{sec-semparse.tex}

	\input{sec-counterfactual.tex}
	\input{sec-corpus.tex}

	\input{sec-experiment.tex}
	\input{sec-conclusion.tex}
	
	\section*{Acknowledgments}
	The research reported in this paper was supported in part by DFG grant RI-2221/4-1.	
	\bibliography{acl2018}
	\bibliographystyle{acl_natbib}
	\clearpage
	\begin{appendix}
		\section*{\LARGE{Supplementary Material}}
		\input{appendix.tex}

	\end{appendix}
	
\end{document}

%% file: sec-intro.tex
\section{Introduction}

\begin{figure*}
	\centering
	\includegraphics[width=1.0\textwidth,keepaspectratio]{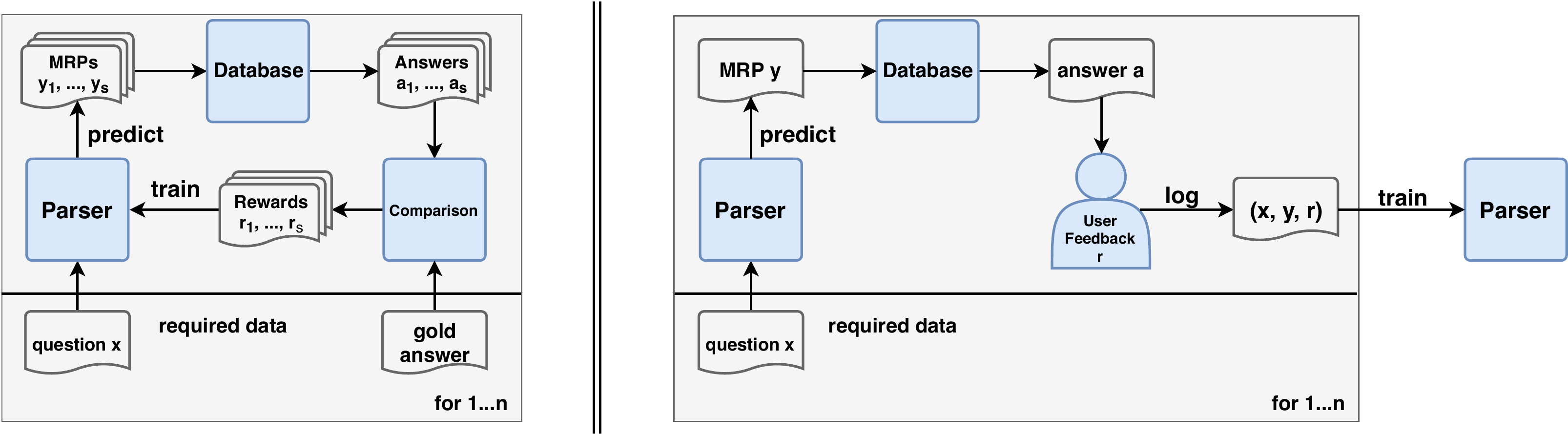}
	\caption{Left: Online reinforcement learning setup for semantic parsing setup where both questions and gold answers are available. The parser attempts to find correct machine readable parses (MRPs) by producing multiple parses, obtaining corresponding answers,  and comparing them against the gold answer.
		Right: In our setup, a question does not have an associated gold answer. The parser outputs a single MRP and the corresponding answer is shown to a user who provides some feedback. Such triplets are collected in a log which can be used for offline training of a semantic parser. This scenario is called counterfactual since the feedback was logged for outputs from a system different from the target system to be optimized.}
	\label{fig:diagram}
\end{figure*}

In semantic parsing, natural language utterances are mapped to machine readable parses which are complex and often tailored specifically to the underlying task. The cost and difficulty of manually preparing large amounts of such parses thus is a bottleneck for supervised learning in semantic parsing.
Recent work (\citet{LiangETAL:17,MouETAL:17a,PengETAL:17}; \textit{inter alia}) has applied reinforcement learning to address the annotation bottleneck as follows: Given a question, the existence of a corresponding gold answer is assumed. A semantic parser produces multiple parses per question and corresponding answers are obtained. These answers are then compared against the gold answer and a positive reward is recorded if there is an overlap. The parser is then guided towards correct parses using the REINFORCE algorithm \citep{Williams:92} which scales the gradient for the various parses by their obtained reward (see the left half of Figure \ref{fig:diagram}). However, learning from question-answer pairs is only efficient if gold answers are cheap to obtain. For complex open-domain question-answering tasks, correct answers are not unique factoids, but open-ended lists, counts in large ranges, or fuzzily defined objects. For example, geographical queries against databases such as OpenStreetMap (OSM) can involve fuzzy operators such as ``near'' or ``in walking distance'' and thus need to allow for fuzziness in the answers as well. 
A possible solution lies in machine learning from even weaker supervision signals in form of human bandit feedback\footnote{The term ``bandit feedback'' is inspired by the scenario of maximizing the reward for a sequence of pulls of arms of ``one-armed bandit'' slot machines.} where the semantic parsing system suggests exactly one parse for which feedback is collected from a human user. In this setup neither gold parse nor gold answer are known and feedback is obtained for only one system output per question.

The goal of our paper is to exploit this scenario of learning from human bandit feedback to train semantic parsers. This learning scenario perfectly fits commercial setups such as virtual personal assistants that embed a semantic parser. Commercial systems can easily log large amounts of interaction data between users and system.
Once sufficient data has been collected, the log can then be used to improve the parser. This leads to a counterfactual learning scenario \cite{BottouETAL:13} where we have to solve the counterfactual problem of how to improve a target system from logged feedback that was given to the outputs of a different historic system (see the right half of Figure \ref{fig:diagram}).

In order to achieve our goal of counterfactual learning of semantic parsers from human bandit feedback, the following contributions are required: First, we need to construct an easy-to-use user interface that allows to collect feedback based on the parse rather than the answer. To this aim, we automatically convert the parse to a set of statements that can be judged as correct or incorrect by a human. This approach allows us to assign rewards at the token level, which in turn enables us to perform blame assignment in bandit learning and to learn from partially correct queries where tokens are reinforced individually. We show that users can provide such feedback for one question-parse pair in 16.4 seconds on average. This exemplifies that our approach is more efficient and cheaper than recruiting experts to annotate parses or asking workers to compile large answer sets.

Next, we demonstrate experimentally that counterfactual learning can be applied to neural sequence-to-sequence learning for semantic parsing. A baseline neural semantic parser is trained in fully supervised fashion, human bandit feedback from human users is collected in a log and subsequently used to improve the parser. The resulting parser significantly outperforms the baseline model as well as a simple bandit-to-supervised approach (B2S) where the subset of completely correct parses is treated as a supervised dataset. Finally, we repeat our experiments on a larger but simulated log to show that our gains can scale: the baseline system is improved by 7.45\% in answer F1 score without ever seeing a gold standard parse.

Lastly, from a machine learning perspective, we have to solve problems of degenerate behavior in counterfactual learning by lifting the multiplicative control variate technique \cite{SwaminathanJoachimsNIPS:15,LawrenceETAL:17,LawrenceETALnips:17} to stochastic learning for neural models. This is done by reweighting target model probabilities over the logged data  under a one-step-late model that decouples the normalization from gradient estimation and is thus applicable in stochastic (minibatch) gradient optimization.

%% file: sec-related.tex
\section{Related Work}
Semantic parsers have been successfully trained using neural sequence-to-sequence models with a cross-entropy objective and question-parse pairs \cite{JiaLiang:16,DongLapata:16}) or question-answer pairs \cite{NeelakantanETAL:16}. Improving semantic parsers using weak feedback has previously been studied (\citet{GoldwasserRoth:13}; \citet{ArtziZettlemoyer:13}; \textit{inter alia}). More recently, several works have applied policy gradient techniques such as REINFORCE \citep{Williams:92} to train neural semantic parsers (\citet{LiangETAL:17}; \citet{MouETAL:17a}; \citet{PengETAL:17}; \textit{inter alia}). However, they assume the existence of the true target answers that can be used to obtain a reward for any number of output queries suggested by the system. It thus differs from a bandit setup where we assume that a reward is available for only one structure.

Our work most closely resembles the work of \citet{IyerETAL:17} who do make the assumption of only being able to judge one output. They improve their parser using simulated and real user feedback. Parses with negative feedback are given to experts to obtain the correct parse. Corrected queries and queries with positive feedback are added to the training corpus and learning continues with a cross-entropy objective. 
We show that this bandit-to-supervision approach can be outperformed by offline bandit learning from partially correct queries.
\citet{YihETAL:16} proposed a user interface for the Freebase database that enables a fast and easy creation of parses. However, in their setup the worker still requires expert knowledge about the Freebase database, whereas in our approach feedback can be collected freely and from any user interacting with the system.

From a machine learning perspective, related work can be found in the areas of counterfactual bandit learning \cite{DudikETAL:11,SwaminathanJoachimsJMLR:15}, or equivalently, off-policy reinforcement learning \cite{PrecupETAL:00,JiangLi:16}. Our contribution is to modify the self-normalizing estimator \citep{Kong:92,PrecupETAL:00,SwaminathanJoachimsNIPS:15,JoachimsETAL:18} to be applicable to neural networks.
Our work is similar to the counterfactual learning setup for machine translation introduced by \citet{LawrenceETAL:17}. Following their insight, we also assume the logs were created deterministically, i.e. the logging policy always outputs the most likely sequence. Their framework was applied to statistical machine translation using linear models. We show how to generalize their framework to neural models and how to apply it to the task of neural semantic parsing in the OSM domain. 

%% file: sec-semparse.tex
\section{Neural Semantic Parsing}\label{sec:semparse}

Our semantic parsing model is a state-of-the-art sequence-to-sequence neural network using an encoder-decoder setup \cite{ChoETAL:14,SutskeverETAL:14} together with an attention mechanism \cite{BahdanauETAL:15}. We use the settings of \citet{SennrichETAL:17}, where an input sequence $x = x_1, x_2, \dots x_{|x|}$ (a natural language question) is encoded by a Recurrent Neural Network (RNN), each input token has an associated hidden vector $h_i = [\overrightarrow{h}_i; \overleftarrow{h}_i]$ where the former is created by a forward pass over the input, and the latter by a backward pass. $\overrightarrow{h}_i$ is obtained by recursively computing $f(x_i, \overrightarrow{h}_{i-1})$ where $f$ is a Gated Recurrent Unit (GRU) \citep{ChungETAL:14}, and $\overleftarrow{h}_i$ is computed analogously. The input sequence is reduced to a single vector $c= g(\{h_1, \dots, h_{|x|}\})$ which serves as the initialization of the decoder RNN. $g$ calculates the average over all vectors $h_i$. At each time step $t$ the decoder state is set by $s_t = q(s_{t-1}, y_{t-1}, c_t)$. $q$ is a conditional GRU with an attention mechanism and $c_t$ is the context vector computed by the attention mechanism. Given an output vocabulary $\mathcal{V}_y$ and the decoder state $s_t = \{s_{_1}, \dots, s_{|\mathcal{V}_y|}\}$, a softmax output layer defines a probability distribution over $\mathcal{V}_y$ and the probability for a token $y_j$ is:

\begin{equation}
\pi_w(y_j = t_o|y_{<j},x) = \frac{\exp(s_{t_o})}{\sum_{v=1}^{|\mathcal{V}_y|} \exp(s_{t_v})}.
\end{equation}
The model $\pi_w$ can be seen as parameterized policy over an action space defined by the target language vocabulary. The probability for a full output sequence $y = y_1, y_2, \dots y_{|y|}$ is defined by 

\begin{equation}
\pi_w(y|x) = \prod_{j=1}^{|y|} \pi_w(y_j|y_{<j},x).
\end{equation}
In our case, output sequences are linearized machine readable parses, called queries in the following. Given supervised data $\D_{sup} = \{(x_t,\bar{y}_t)\}_{t=1}^n$ of question-query pairs, where $\bar{y}_t$ is the true target query for $x_t$, the neural network can be trained using SGD and a cross-entropy (CE) objective:

\begin{equation}
\mathcal{L}_{CE} = - \frac{1}{n} \sum_{t=1}^{n} \sum_{j=1}^{|\bar{y}|} \log \pi_w(\bar{y}_{t,j} | \bar{y}_{t,<j}, x_t),
\end{equation}

where $y_{t,<j} = y_{t,1}, y_{t,2} \dots y_{t,j-1}$.

%% file: sec-counterfactual.tex
\section{Counterfactual Learning from Deterministic Bandit Logs}\label{sec:counterfactual}

\begin{table*}[t]
	\begin{center}
		\begin{tabular}{l}
			\toprule
			$\nabla_w \hat{R}_{\text{DPM}} =  \frac{1}{n} \sum_{t=1}^{n} \delta_t \pi_w(y_t | x_t) \nabla_w \log \pi_w(y_t | x_t).$ \\ \midrule
			$\nabla_w\hat{R}_{\text{DPM+R}} = \frac{1}{n}\sum_{t=1}^{n} [ \delta_t \bar{\pi}_w(y_t|x_t) (  \nabla_w \log \pi_w(y_t | x_t)  - \frac{1}{n}\sum_{u=1}^{n} \bar{\pi}_w(y_u|x_u)  \nabla \log \pi_w(y_u | x_u) )].$ \\ \midrule
			$\nabla_w \hat{R}_{\text{DPM+OSL}} = \frac{1}{m}\sum_{t=1}^{m}  \delta_t \bar{\pi}_{w,w'}(y_t|x_t) \nabla_w \log \pi_w(y_t | x_t).$ \\ \midrule
			$\nabla_w \hat{R}_{\text{DPM+T}} =  \frac{1}{n} \sum_{t=1}^{n} \left(\sum_{j=1}^{|y_t|} \delta_{t,j} \nabla_w \log \pi_w(y_{t,j}|y_{t,<j},x_t)\right).$ \\ \midrule
			$\nabla_w \hat{R}_{\text{DPM+T+OSL}} = \frac{\frac{1}{m}\sum_{t=1}^{m}  \left(\sum_{j=1}^{|y_t|} \delta_{t,j} \nabla_w  \log \pi_w(y_{t,j}|y_{t,<j},x_t)\right)}{\frac{1}{n}\sum_{t=1}^{n} \pi_{w'}(y_t|x_t)}.$ \\ 
			\bottomrule
		\end{tabular}
	\end{center}
	\caption{Gradients of counterfactual objectives.}
	\label{tab:gradients}
\end{table*}

\paragraph{Counterfactual Learning Objectives.} We assume a policy $\pi_w$ that, given an input $x \in \mathcal{X}$, defines a conditional probability distribution over possible outputs $y \in \mathcal{Y}(x)$. Furthermore, we assume that the policy is parameterized by $w$ and its gradient can be derived. In this work, $\pi_w$ is defined by the sequence-to-sequence model described in Section \ref{sec:semparse}. We also assume that the model decomposes over individual output tokens, i.e. that the model produces the output token by token.

The counterfactual learning problem can be described as follows: We are given a data log of triples $\D_{log}=\{(x_t,y_t,\delta_t)\}_{t=1}^n$ where outputs $y_t$ for inputs $x_t$ were generated by a logging system under policy $\pi_0$, and loss values $\delta_t \in [-1,0]$\footnote{We use the terms loss and (negative) rewards interchangeably, depending on context.} were observed for the generated data points. Our goal is to optimize the expected reward (in our case: minimize the expected risk) for a target policy $\pi_w$ given the data log $\D_{log}$.
In case of deterministic logging, outputs are logged with propensity $\pi_0(y_t|x_t) = 1, \; t=1, \dots, n$. This results in a \emph{deterministic propensity matching (DPM)} objective \cite{LawrenceETAL:17}, without the possibility to correct the sampling bias of the logging policy by inverse propensity scoring \cite{RosenbaumRubin:83}:
\begin{align}
\label{eq:emp-risk} 
\hat{R}_{\text{DPM}}(\pi_w) = \frac{1}{n} \sum_{t=1}^n \delta_t \pi_w(y_t|x_t).
\end{align}

This objective can show degenerate behavior in that it overfits to the choices of the logging policy \cite{SwaminathanJoachimsNIPS:15,LawrenceETALnips:17}. This degenerate behavior can be avoided by reweighting using a multiplicative control variate \citep{Kong:92,PrecupETAL:00,JiangLi:16,ThomasBrunskill:16}. The new objective is called the \emph{reweighted deterministic propensity matching (DPM+R)} objective in \citet{LawrenceETAL:17}:
\begin{align}
\label{eq:r} 
\hat{R}_{\text{DPM+R}}(\pi_w) &= \frac{1}{n} \sum_{t=1}^{n} \delta_t \bar{\pi}_w(y_t|x_t) \\ \notag
&= \frac{\frac{1}{n}\sum_{t=1}^{n} \delta_t \pi_w(y_t|x_t)}{\frac{1}{n}\sum_{t=1}^{n} \pi_w(y_t|x_t)}.
\end{align}
Algorithms for optimizing the discussed objectives can be derived as gradient descent algorithms where gradients using the score function gradient estimator \citep{Fu:06} are shown in Table \ref{tab:gradients}.

\paragraph{Reweighting in Stochastic Learning.} As shown in \citet{SwaminathanJoachimsNIPS:15} and \citet{LawrenceETALnips:17}, reweighting over the entire data log $\D_{log}$ is crucial since it avoids that high loss outputs in the log take away probability mass from low loss outputs. This multiplicative control variate has the additional effect of reducing the variance of the estimator, at the cost of introducing a bias of order $O(\frac{1}{n})$ that decreases as $n$ increases \cite{Kong:92}. The desirable properties of this control variate cannot be realized in a stochastic (minibatch) learning setup since minibatch sizes large enough to retain the desirable reweighting properties are infeasible for large neural networks.

We offer a simple solution to this problem that nonetheless retains all desired properties of the reweighting. The idea is inspired by one-step-late algorithms that have been introduced for EM algorithms \cite{Green:90}. In the EM case, dependencies in objectives are decoupled by evaluating certain terms under parameter settings from previous iterations (thus: one-step-late) in order to achieve closed-form solutions. In our case, we decouple the reweighting from the parameterization of the objective by evaluating the reweighting under parameters $w'$ from some previous iteration. This allows us to perform gradient descent updates and reweighting asynchronously. Updates are performed using minibatches, however, reweighting is based on the entire log, allowing us to retain the desirable properties of the control variate.

The new objective, called \textit{one-step-late reweighted DPM} objective (DPM+OSL), optimizes $\pi_{w,w'}$ with respect to $w$ for a minibatch of size $m$, with reweighting over the entire log of size $n$ under parameters $w'$:
\begin{align}
\label{eq:prob_mini} 
\hat{R}_{\text{DPM+OSL}}(\pi_w) &= \frac{1}{m}\sum_{t=1}^{m} \delta_t \bar{\pi}_{w,w'}(y_t|x_t) \\ \notag
&= \frac{\frac{1}{m}\sum_{t=1}^{m} \delta_t \pi_w(y_t|x_t)}{\frac{1}{n}\sum_{t=1}^{n} \pi_{w'}(y_t|x_t)}.
\end{align}
If the renormalization is updated periodically, e.g. after every validation step, renormalizations under $w$ or $w'$ are not much different and will not hamper convergence. Despite losing the formal justification from the perspective of control variates, we found empirically that the OSL update schedule for reweighting is sufficient and does not deteriorate performance. The gradient for learning with OSL updates is given in Table \ref{tab:gradients}.

\paragraph{Token-Level Rewards.} For our application of counterfactual learning to human bandit feedback, we found another deviation from standard counterfactual learning to be helpful: For humans, it is hard to assign a graded reward to a query at a sequence level because either the query is correct or it is not. In particular, with a sequence level reward of 0 for incorrect queries, we do not know which part of the query is wrong and which parts might be correct. Assigning rewards at token-level will ease the feedback task and allow the semantic parser to learn from partially correct queries. Thus, assuming the underlying policy can decompose over tokens, a token level (DPM+T) reward objective can be defined:
\begin{align}
\label{eq:token} 
&\hat{R}_{\text{DPM+T}}(\pi_w) = \\ \notag
&\frac{1}{n} \sum_{t=1}^{n} \left(\sum_{j=1}^{|y|} \delta_{t,j} \log \pi_w(y_{t,j}|y_{t,<j}, x_t)\right).
\end{align}
Analogously, we can define an objective that combines the token-level rewards and the minibatched reweighting (DPM+T+OSL):
\begin{align}
\label{eq:token_prob} 
&\hat{R}_{\text{DPM+T+OSL}}(\pi_w) = \\ \notag
&\frac{\frac{1}{m}\sum_{t=1}^{m}  \left(\sum_{j=1}^{|y|} \delta_{t,j} \log \pi_w(y_{t,j}|y_{t,<j}, x_t)\right)}{\frac{1}{n}\sum_{t=1}^{n} \pi_{w'}(y_t|x_t)}.
\end{align}
Gradients for the DPM+T and DPM+T+OSL objectives are given in Table \ref{tab:gradients}.

%% file: sec-corpus.tex
\section{Semantic Parsing in the OpenStreetMap Domain}\label{sec:nlmaps}

OpenStreetMap (OSM) is a geographical database in which volunteers annotate points of interests in the world. A point of interest consists of one or more associated GPS points. Further relevant information may be added at the discretion of the volunteer in the form of tags. Each tag consists of a key and an associated value, for example ``\textit{tourism : hotel}''.
The \textsc{NLmaps} corpus was introduced by \citet{HaasRiezler:16} as a basis to create a natural language interface to the OSM database. It pairs English questions with machine readable parses, i.e. queries that can be executed against OSM.

\begin{figure}
	\centering
	\includegraphics[width=0.45\textwidth,keepaspectratio]{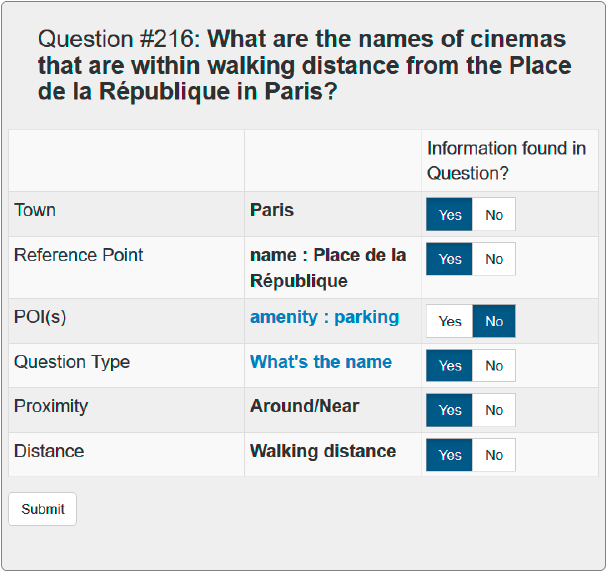}
	\caption{The user interface for collecting feedback from humans with an example question and a correctly filled out form.}
	\label{fig:user_interface}
\end{figure}

\paragraph{Human Feedback Collection.}
The task of creating a natural language interface for OSM demonstrates typical difficulties that make it expensive to collect supervised data. The machine readable language of the queries is based on the \textsc{Overpass} query language which was specifically designed for the OSM database. It is thus not easily possible to find experts that could provide correct queries. It is equally difficult to ask workers at crowdsourcing platforms for the correct answer. For many questions, the answer set is too large to expect a worker to count or list them all in a reasonable amount of time and without errors. For example, for the question ``\textit{How many hotels are there in Paris?}'' there are 951 hotels annotated in the OSM database.
Instead we propose to automatically transform the query into a block of statements that can easily be judged as correct or incorrect by a human. The question and the created block of statements are embedded in a user interface with a form that can be filled out by users. Each statement is accompanied by a set of radio buttons where a user can select either ``\textit{Yes}'' or ``\textit{No}''. For a screenshot of the interface and an example see Figure \ref{fig:user_interface}.

In total there are 8 different types of statements.
The presence of certain tokens in a query trigger different statement types. For example, the token ``\textit{area}'' triggers the statement type ``\textit{Town}''. The statement is then populated with the corresponding information from the query. In the case of ``\textit{area}'', the following OSM value is used, e.g. ``\textit{Paris}''. With this, the meaning of every query can be captured by a set of human-understandable statements. For a full overview of all statement types and their triggers see section \ref{supp:feedback} of the supplementary material.

OSM tags and keys are generally understandable. For example, the correct OSM tag for ``\textit{hotels}'' is ``\textit{tourism : hotel}'' and when searching for websites, the correct question type key would be ``\textit{website}''. Nevertheless, for each OSM tag or key, we automatically search for the corresponding Wikipedia page on the OpenStreetMap Wiki\footnote{\url{https://wiki.openstreetmap.org/}} and extract the description for this tag or key. The description is made available to the user in form of a tool-tip that appears when hovering over the tag or key with the mouse. If a user is unsure if a OSM tag or key is correct, they can read this description to help in their decision making. Once the form is submitted, a script maps each statement back to the corresponding tokens in the original query. These tokens then receive negative or positive feedback based on the feedback the user provided for that statement.

\paragraph{Corpus Extension.} Similar to the extension of the \textsc{NLmaps} corpus by \citet{LawrenceRiezler:16} who include shortened questions which are more typically used by humans in search tasks, we present an automatic extension that allows a larger coverage of common OSM tags.\footnote{The extended dataset, called \textsc{NLmaps v2}, will be released upon acceptance of the paper.}
The basis for the extension is a hand-written, online freely available list\footnote{\url{http://wiki.openstreetmap.org/wiki/Nominatim/Special\_Phrases/EN}} that links natural language expressions such as ``\textit{cash machine}'' to appropriate OSM tags, in this case ``\textit{amenity : atm}''.
Using the list, we generate for each unique expression-tag pair a set of question-query pairs. These latter pairs contain several placeholders which will be filled automatically in a second step.

To fill the area placeholder \texttt{\$LOC}, we sample from a list of 30 cities from France, Germany and the UK. \texttt{\$POI} is the placeholder for a point of interest. We sample it from the list of objects which are located in the prior sampled city and which have a \textit{name} key. The corresponding value belonging to the \textit{name} key will be used to fill this spot. The placeholder \texttt{\$QTYPE} is filled by uniformly sampling from the four primary question types available in the \textsc{NLmaps} query language. On the natural language side they corresponded to ``\textit{How many}'', ``\textit{Where}'', ``\textit{Is there}'' and \texttt{\$KEY}. \texttt{\$KEY} is a further parameter belonging to the primary question operator \textsc{findkey}. It can be filled by any OSM key, such as \textit{name}, \textit{website} or \textit{height}. To ensure that there will be an answer for the generated query, we first ran a query with the current tag (``\textit{amenity : atm}'') to find all objects fulfilling this requirement in the area of the already sampled city. From the list of returned objects and the keys that appear in association with them, we uniformly sampled a key. For \texttt{\$DIST} we chose between the pre-defined options for walking distance and within city distance. The expressions map to corresponding values which define the size of a radius in which objects of interest (with tag ``\textit{amenity : atm}'') will be located. If the walking distance was selected, we added ``\textit{in walking distance}'' to the question. Otherwise no extra text was added to the question, assuming the within city distance to be the default. This sampling process was repeated twice. 

\begin{table}[t]
	\centering
	\tabcolsep=0.11cm
	\begin{tabular}{lll}
		\toprule
		& \textsc{NLmaps} & \textsc{NLmaps v2}\\
		\midrule
		\# question-query pairs & 2,380 & 28,609 \\
		tokens & 25,906 & 202,088 \\
		types & 1,002 & 8,710 \\
		avg. sent. length & 10.88 & 7.06 \\
		distinct tags & 477 & 6,582 \\
		\bottomrule
	\end{tabular}
	\caption{Corpus statistics of the question-answering corpora \textsc{NLmaps} and our extension \textsc{NLmaps v2} which additionally contains the search engine style queries \citep{LawrenceRiezler:16} and the automatic extensions of the most common OSM tags.}
	\label{tab:nlmaps_statistics2}
\end{table}

Table \ref{tab:nlmaps_statistics2} presents the corpus statistics, comparing \textsc{NLmaps} to our extension. The automatic extension, obviating the need for expensive manual work, allows a vast increase of question-query pairs by an order of magnitude. Consequently the number of tokens and types increase in a similar vein. However, the average sentence length drops. This comes as no surprise due to the nature of the rather simple hand-written list which contains never more than one tag for an element, resulting in simpler question structures. However, the main idea of utilizing this list is to extend the coverage to previously unknown OSM tags. With 6,582 distinct tags compared to the previous 477, this was clearly successful. Together with the still complex sentences from the original corpus, a semantic parser is now able to learn both complex questions and a large variety of tags. An experiment that empirically validates the usefulness of the automatically created data can be found in the supplementary material, section \ref{supp:nlmaps}.

%% file: sec-experiment.tex
\section{Experiments}\label{sec:exp}

\paragraph{General Settings.}
In our experiments we use the sequence-to-sequence neural network package \textsc{Nematus} \citep{SennrichETAL:17}. Following the method used by \citet{HaasRiezler:16}, we split the queries into individual tokens by taking a pre-order traversal of the original tree-like structure. For example, ``\textit{query(west(area(keyval('name','Paris')), nwr(keyval('railway','station'))),qtype(count))}'' becomes ``\textit{query@2 west@2 area@1 keyval@2 name@0 Paris@s nwr@1 keyval@2 railway@0 station@s qtype@1 count@0}''.

The SGD optimizer used is ADADELTA \citep{Zeiler:12}. The model employs 1,024 hidden units and word embeddings of size 1,000. The maximum sentence length is 200 and gradients are clipped if they exceed a value of 1.0. The stopping point is determined by validation on the development set and selecting the point at which the highest evaluation score is obtained. F1 validation is run after every 100 updates, and each update is made on the basis of a minibatch of size 80.

The evaluation of all models is based on the answers obtained by executing the most likely query obtained after a beam search with a beam of size 12. We report the F1 score which is the harmonic mean of precision and recall. Recall is defined as the percentage of fully correct answers divided by the set size. Precision is the percentage of correct answers out of the set of answers with non-empty strings.  Statistical significance between models is measured using an approximate randomization test \citep{Noreen:1989}.

\paragraph{Baseline Parser \& Log Creation.} Our experiment design assumes a baseline neural semantic parser that is trained in fully supervised fashion, and is to be improved by bandit feedback obtained for system outputs from the baseline system for given questions. For this purpose, we select 2,000 question-query pairs randomly from the full extended \textsc{NLmaps v2} corpus. We will call this dataset $\D_{sup}$. Using this dataset, a baseline semantic parser is trained in supervised fashion under a cross-entropy objective. It obtains an F1 score of 57.45\% and serves as the logging policy $\pi_0$.

Furthermore we randomly split off 1,843 and 2,000 pairs for a development and test set, respectively. This leaves a set of 22,765 question-query pairs. The questions can be used as input and bandit feedback can be collected for the most likely output of the semantic parser. We refer to this dataset as $\D_{log}$.

To collect human feedback, we take the first 1,000 questions from $\D_{log}$ and use $\pi_0$ to parse these questions to obtain one output query for each. 5 question-query pairs are discarded because the suggested query is invalid. For the remaining question-query pairs, the queries are each transformed into a block of human-understandable statements and embedded into the user interface described in Section \ref{sec:nlmaps}.
We recruited 9 users to provide feedback for these question-query pairs. The resulting log is referred to as $\D_{human}$. Every question-query pair is purposely evaluated only once to mimic a realistic real-world scenario where user logs are collected as users use the system. In this scenario, it is also not possible to explicitly obtain several evaluations for the same question-query pair. Some examples of the  received feedback can be found in the supplementary material, section \ref{supp:screenshots}.

To verify that the feedback collection is efficient, we measured the time each user took from loading a form to submitting it. To provide feedback for one question-query pair, users took 16.4 seconds on average with a standard deviation of 33.2 seconds. The vast majority (728 instances) are completed in less than 10 seconds.

\paragraph{Learning from Human Bandit Feedback.}
An analysis of $\D_{human}$ shows that for 531 queries all corresponding statements were marked as correct. We consider a simple baseline that treats completely correct logged data as a supervised data set with which training continues using the cross-entropy objective. We call this baseline bandit-to-supervised conversion (B2S).
Furthermore, we present experimental results using the log $\D_{human}$ for stochastic (minibatch) gradient descent optimization of the counterfactual objectives introduced in equations \ref{eq:emp-risk}, \ref{eq:prob_mini}, \ref{eq:token} and \ref{eq:token_prob}. For the token-level feedback, we map the evaluated statements back to the corresponding tokens in the original query and assign these tokens a feedback of 0 if the corresponding statement was marked as wrong and 1 otherwise. In the case of sequence-level feedback, the query receives a feedback of 1 if all statements are marked correct, 0 otherwise. For the OSL objectives, a separate experiment (see below) showed that updating the reweighting constant after every validation step promises the best trade-off between performance and speed.

Results, averaged over 3 runs, are reported in Table \ref{exp:human_results}. The B2S model can slightly improve upon the baseline but not significantly. DPM improves further, significantly beating the baseline. Using the multiplicative control variate modified for SGD by OSL updates does not seem to help in this setup.
By moving to token-level rewards, it is possible to learn from partially correct queries. These partially correct queries provide valuable information that is not present in the subset of correct answers employed by the previous models. Optimizing DPM+T leads to a slight improvement and combined with the multiplicative control variate, DPM+T+OSL yields an improvement of about 1.0 in F1 score upon the baseline. It beats both the baseline and the B2S model by a significant margin. 

\begin{table}
	\begin{center}
		\begin{tabular}{llll}
			\toprule
			&& F1 & $\Delta$ F1 \\
			\midrule
			\raisebox{.1\height}{\scriptsize 1}&baseline&57.45&\\
			\raisebox{.1\height}{\scriptsize 2}&B2S&57.79\small$\pm0.18$&+0.34\\
			\raisebox{.1\height}{\scriptsize 3}&DPM$^1$&58.04\small$\pm0.04$&+0.59\\
			\raisebox{.1\height}{\scriptsize 4}&DPM+OSL&58.01\small$\pm0.23$&+0.56\\
			\raisebox{.1\height}{\scriptsize 5}&DPM+T$^1$&58.11\small$\pm0.24$&+0.66\\
			\raisebox{.1\height}{\scriptsize 6}&DPM+T+OSL$^{1,2}$&58.44\small$\pm0.09$&+0.99\\
			\bottomrule
		\end{tabular}
		\caption{Human Feedback: Answer F1 scores on the test set for the various setups, averaged over 3 runs. Statistical significance of system differences at $p<0.05$ are indicated by experiment number in superscript.}
		\label{exp:human_results}
	\end{center}
\end{table}

\paragraph{Learning from Large-Scale Simulated Feedback.}
We want to investigate whether the results scale if a larger log is used. Thus, we use $\pi_0$ to parse all 22,765 questions from $\D_{log}$ and obtain for each an output query. For sequence level rewards, we assign feedback of $1$ for a query if it is identical to the true target query, 0 otherwise. We also simulate token-level rewards by iterating over the indices of the output and assigning a feedback of $1$ if the same token appears at the current index for the true target query, 0 otherwise.

An analysis of $\D_{log}$ shows that 46.27\% of the queries have a sequence level reward of $1$ and are thus completely correct. This subset is used to train a bandit-to-supervised (B2S) model using the cross-entropy objective.

Experimental results for the various optimization setups, averaged over 3 runs, are reported in Table \ref{exp:results}.
We see that the B2S model outperforms the baseline model by a large margin, yielding an increase in F1 score by 6.24 points. Optimizing the DPM objective also yields a significant increase over the baseline, but its performance falls short of the stronger B2S baseline. Optimizing the DPM+OSL objective leads to a substantial improvement in F1 score over optimizing DPM but still falls slightly short of the strong B2S baseline.
Token-level rewards are again crucial to beat the B2S baseline significantly. DPM+T is already able to significantly outperform B2S in this setup and DPM+T+OSL can improve upon this further.

\begin{table}
	\begin{center}
		\begin{tabular}{llll}
			\toprule
			&& F1 & $\Delta$ F1 \\
			\midrule
			\raisebox{.1\height}{\scriptsize 1}&baseline&57.45&\\
			\raisebox{.1\height}{\scriptsize 2}&B2S$^{1,3}$&63.22\small$\pm0.27$&+5.77\\
			\raisebox{.1\height}{\scriptsize 3}&DPM$^{1}$&61.80\small$\pm0.16$&+4.35\\
			\raisebox{.1\height}{\scriptsize 4}&DPM+OSL$^{1,3}$&62.91\small$\pm0.05$&+5.46\\
			\raisebox{.1\height}{\scriptsize 5}&DPM+T$^{1,2,3,4}$&63.85\small$\pm0.2$&+6.40\\
			\raisebox{.1\height}{\scriptsize 6}&DPM+T+OSL$^{1,2,3,4}$&64.41\small$\pm0.05$&+6.96\\
			\bottomrule
		\end{tabular}
		\caption{Simulated Feedback: Answer F1 scores on the test set for the various setups, averaged over 3 runs. Statistical significance of system differences at $p<0.05$ are indicated by experiment number in superscript.}
		\label{exp:results}
	\end{center}
\end{table}

\paragraph{Analysis.}
Comparing the baseline and DPM+T+OSL, we manually examined all queries in the test set where DPM+T+OSL obtained the correct answer and the baseline system did not (see Table \ref{exp:analysis}). The analysis showed that the vast majority of previously wrong queries were fixed by correcting an OSM tag in the query. For example, for the question ``\textit{closest Florist from Manchester in walking distance}'' the baseline system chose the tag ``\textit{landuse : retail}'' in the query, whereas DPM+T+OSL learnt that the correct tag is ``\textit{shop : florist}''. In some cases, the question type had to be corrected, e.g. the baseline's suggested query returned the location of a point of interest but DPM+T+OSL correctly returns the phone number. Finally, in a few cases DPM+T+OSL corrected the structure for a query, e.g. by searching for a point of interest in the east of an area rather than the south.

\begin{table}
	\begin{center}
		\begin{tabular}{lll}
			\toprule
			Error Type& Human & Simulated \\
			\midrule
			OSM Tag&90\%&86.75\%\\
			Question Type&6\%&8.43\%\\
			Structure&4\%&4.82\%\\
			\bottomrule
		\end{tabular}
		\caption{Analysis of which type of errors DPM+T+OSL corrected on the test set compared to the baseline system for both human and simulated feedback experiments.}
		\label{exp:analysis}
	\end{center}
\end{table}

\paragraph{OSL Update Variation.}\label{sec:osl}
Using the DPM+T+OSL objective and the simulated feedback setup, we vary the frequency of updating the reweighting constant. Results are reported in Table \ref{exp:osl_vary_updates}. Calculating the constant only once at the beginning leads to a near identical result in F1 score as not using OSL. The more frequent update strategies, once or four times per epoch, are more effective. Both strategies reduce variance further and lead to higher F1 scores. Updating four times per epoch compared to once per epoch, leads to a nominally higher performance in F1. It has the additional benefit that the re-calculation is done at the same time as the validation, leading to no additional slow down as executing the queries for the development set against the database takes longer than the re-calculation of the constant. Updating after every minibatch is infeasible as it slows down training too much. Compared to the previous setup, iterating over one epoch takes approximately an additional 5.5 hours.

\begin{table}[h]
	\begin{center}
		\begin{tabular}{llll}
			\toprule
			&OSL Update& F1 & $\Delta$ F1 \\
			\midrule\raisebox{.1\height}{\scriptsize 1}&no OSL (DPM+T)&63.85\small$\pm0.2$&\\
			\raisebox{.1\height}{\scriptsize 2}&once&63.82\small$\pm0.1$&-0.03\\
			\raisebox{.1\height}{\scriptsize 3}&every epoch&64.26\small$\pm0.04$&+0.41\\
			\raisebox{.1\height}{\scriptsize 4}&every validation / &\multirow{2}{*}{64.41\small$\pm0.05$}&\multirow{2}{*}{+0.56}\\
			& 4x per epoch &&\\
			\raisebox{.1\height}{\scriptsize 5}&every minibatch&N/A&N/A\\
			\bottomrule
		\end{tabular}
		\caption{Simulated Feedback: Answer F1 scores on the test set for DPM+T and DPM+T+OSL with varying OSL update strategies, averaged over 3 runs. Updating after every minibatch is infeasible as it significantly slows down learning. Statistical significance of system differences at $p<0.05$ occur for experiment 4 over experiment 2.}
		\label{exp:osl_vary_updates}
	\end{center}
\end{table}

%% file: sec-conclusion.tex
\section{Conclusion}

We introduced a scenario for improving a neural semantic parser from logged bandit feedback. This scenario is important to avoid complex and costly data annotation for supervise learning, and it is realistic in commercial applications where weak feedback can be collected easily in large amounts from users. We presented robust counterfactual learning objectives that allow to perform stochastic gradient optimization which is crucial in working with neural networks. 
Furthermore, we showed that it is essential to obtain reward signals at the token-level in order to learn from partially correct queries. We presented experimental results using feedback collected from humans and a larger scale setup with simulated feedback. In both cases we show that a strong baseline using a bandit-to-supervised conversion can be significantly outperformed by a combination of a one-step-late reweighting and token-level rewards.
Finally, our approach to collecting feedback can also be transferred to other domains. For example, \cite{YihETAL:16} designed a user interface to help Freebase experts to efficiently create queries. This interface could be reversed: given a question and a query produced by a parser, the interface is filled out automatically and the user has to verify if the information fits.

%% file: appendix.tex
\section{Empirical Validation of the \textsc{NLmaps} Corpus Extension}\label{supp:nlmaps}
To empirically validate the usefulness of the automatically created data, we compare two parsers trained with \textsc{Nematus} \citep{SennrichETAL:17} (see Section 6 for more details). 
The first model is trained using the original \textsc{NLmaps} training data. The second receives an additional 15,000 instances from the synthetic data. Both systems are tested on the original \textsc{NLmaps} test data and on the new test set of \textsc{NLmaps v2} which consists of a random set of 2,000 pairs from the remaining data. Results may be found in Table \ref{tab:v1_v2_comparison}. On the original test set, adding the 15,000 synthetic instances allows the parser to significantly improve by 2.09 in F1 score. The parser trained on the original training data performs badly on the new test set because it is ignorant of many OSM tags that were introduced with the extension.

\begin{table}[h]
	\centering
	\tabcolsep=0.11cm
	\begin{tabular}{lllll}
		\toprule
		&&\multicolumn{2}{c}{Train}&\\
		&& \textsc{v1} & \textsc{v1+15k}&$\Delta$\\
		\multirow{2}{*}{Test}&\textsc{v1}&73.56\small$\pm0.61$&75.65\small$\pm0.34$&+2.09\\
		&\textsc{v2}&28.31\small$\pm0.25$&79.17\small$\pm0.11$&+50.86\\
		\midrule
		\bottomrule
	\end{tabular}
	\caption{Answer F1 scores on the \textsc{NLmaps v1} and \textsc{NLmaps v2} test sets for models trained on either only \textsc{NLmaps v1} training data or with an additional 15k synthetic instances. Results are averaged over 3 runs. Using the \textsc{NLmaps v2} training set leads to significant system differences on both test sets at $p<0.05$.}
	\label{tab:v1_v2_comparison}
\end{table}

\begin{table*}
	\begin{center}
		\begin{tabular}{ll}
			\toprule
			Type& Explanation  \\
			\midrule
			Town & OSM tags of ``\textit{area}''\\
			Reference Point &OSM tags ``\textit{center}''\\
			POI(s)&OSM tags of ``\textit{search}'' if ``\textit{center}'' is set,\\
			& else of ``\textit{nwr}''\\
			Question Type &Arguments of ``\textit{qtype}''\\
			Proximity : Around/Near&If ``\textit{around}'' is present\\
			Restriction : Closest&If ``\textit{around}'' and ``\textit{topx}'' are present\\
			Distance&Argument of ``\textit{maxdist}''\\
			Cardinal Direction&``\textit{north}'', ``\textit{east}'', ``\textit{south}'' or ``\textit{west}'' are present\\
			\bottomrule
		\end{tabular}
		\caption{Overview of the possible statements types that are used to transform a parse into a human-understandable block of statements.}
		\label{exp:statement_types}
	\end{center}
\end{table*}

\section{Automatic Feedback Form Creation}\label{supp:feedback}
A query can be automatically converted into a set of statements which can be judged as correct or incorrect by non-expert users. There are 8 different statement types and each is triggered based on the shape of the query and certain tokens. An overview of the statement types, their triggers and the value a statement will hold, can be found in Table \ref{exp:statement_types}.

\section{Screenshots of Human User Feedback.}\label{supp:screenshots}
\begin{figure*}
	\centering
	\includegraphics[width=0.8\textwidth,keepaspectratio]{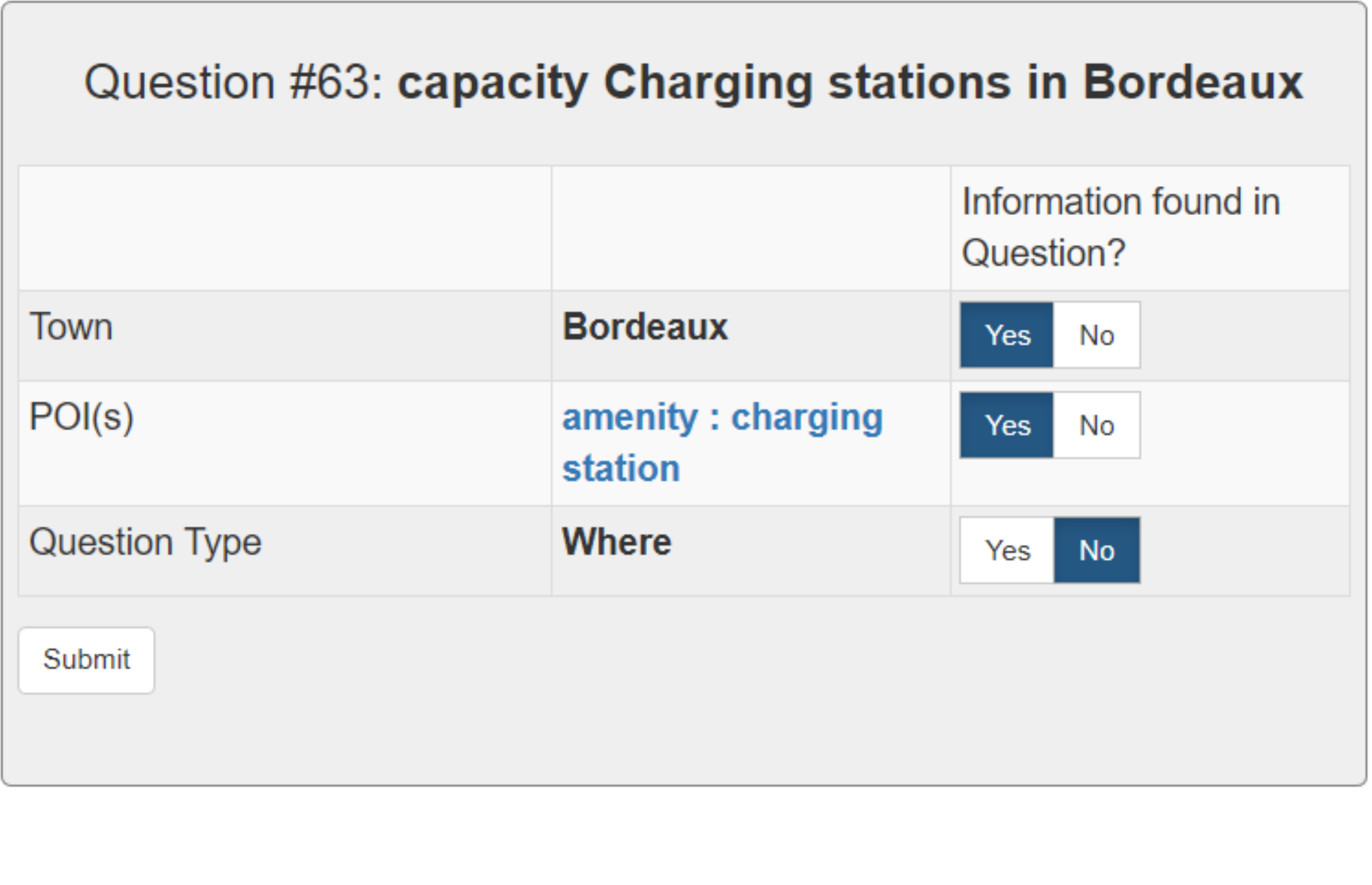}
	\caption{Feedback form for question \#63 as filled out by a human user.}
	\label{fig:user_interface2}
\end{figure*}

\begin{figure*}
	\centering
	\includegraphics[width=0.8\textwidth,keepaspectratio]{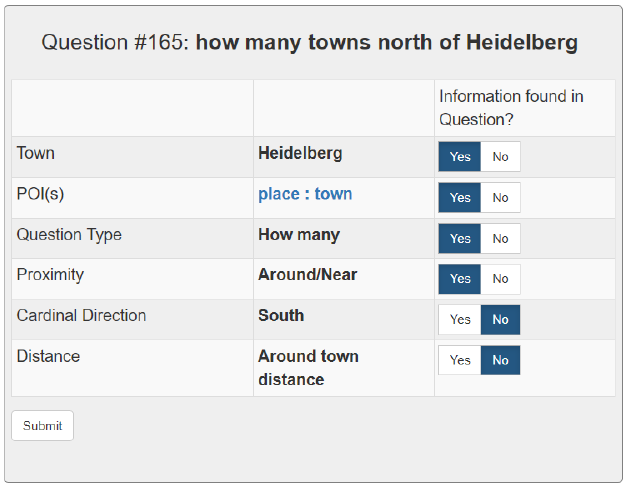}
	\caption{Feedback form for question \#165 as filled out by a human user.}
	\label{fig:user_interface3}
\end{figure*}

\begin{figure*}
	\centering
	\includegraphics[width=0.8\textwidth,keepaspectratio]{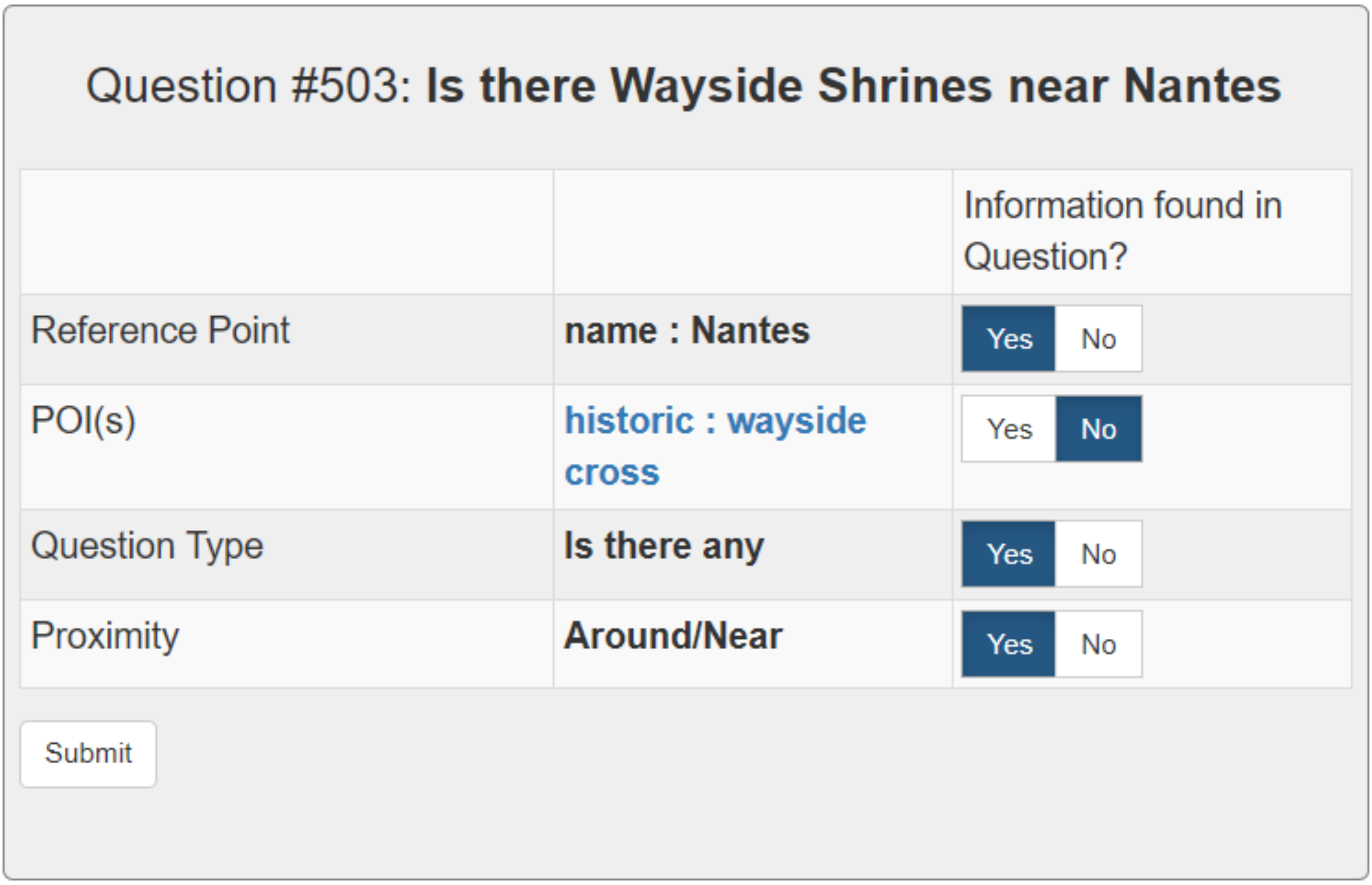}
	\caption{Feedback form for question \#503 as filled out by a human user.}
	\label{fig:user_interface4}
\end{figure*}

\begin{figure*}
	\centering
	\includegraphics[width=0.8\textwidth,keepaspectratio]{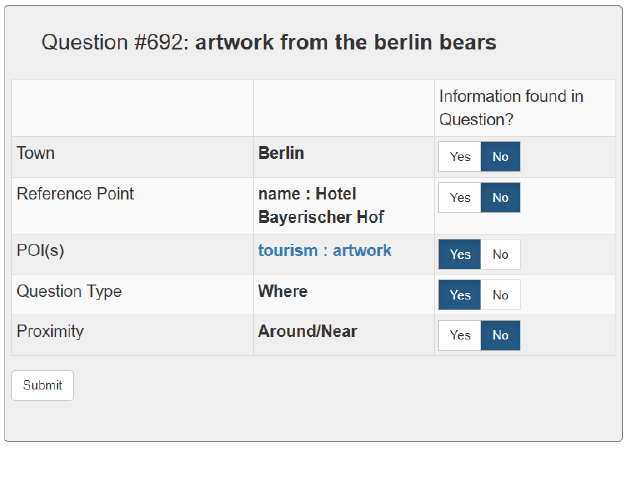}
	\caption{Feedback form for question \#692 as filled out by a human user.}
	\label{fig:user_interface5}
\end{figure*}

\begin{figure*}
	\centering
	\includegraphics[width=0.8\textwidth,keepaspectratio]{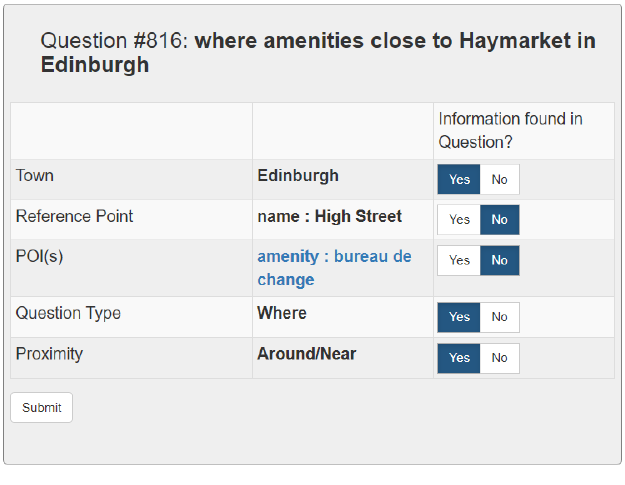}
	\caption{Feedback form for question \#816 as filled out by a human user.}
	\label{fig:user_interface6}
\end{figure*}

Figures \ref{fig:user_interface2}, \ref{fig:user_interface3}, \ref{fig:user_interface4}, \ref{fig:user_interface5} and \ref{fig:user_interface6} present screenshots of forms as filled out by the recruited human users.

